%% file: colm2026_conference.tex
\documentclass{article} 
\usepackage[final]{colm2026_conference}

\usepackage{microtype}
\usepackage{hyperref}
\usepackage{url}
\usepackage{booktabs}
\usepackage[most]{tcolorbox}
\usepackage{xcolor}
\usepackage{makecell}
\usepackage{multirow}
\usepackage{enumitem}
\usepackage{wrapfig}
\usepackage{tabularx}
\usepackage{graphicx}

\usepackage{tikz}
\usetikzlibrary{
    arrows.meta,
    decorations.pathreplacing,
    calc,
    positioning
}

\usepackage{lineno}

\definecolor{darkblue}{rgb}{0, 0, 0.5}
\hypersetup{colorlinks=true, citecolor=darkblue, linkcolor=darkblue, urlcolor=darkblue}

\definecolor{mygreenbg}{RGB}{230,245,235}
\definecolor{mygreenframe}{RGB}{40,120,80}

\newtcolorbox{keyinsight}{
    colback=mygreenbg,
    colframe=mygreenframe,
    boxrule=0.8pt,
    arc=4pt,
    left=8pt,
    right=8pt,
    top=4pt,
    bottom=4pt,
}

\newtcolorbox{hypothesisbox}{
  colback=violet!7,
  colframe=violet!35!black,
  boxrule=0.5pt,
  arc=3pt,
  left=6pt,
  right=6pt,
  top=4pt,
  bottom=4pt
}

\title{LF²AR: Accounting for Layerwise Dynamics to\\Improve Multimodal Adaptation of Language Models}

\author{%
\makebox[\dimexpr\textwidth-2\tabcolsep\relax][c]{%
\begin{tabular}{@{}cccc@{}}
Santiago Cuervo\textsuperscript{1} &
Adel Moumen\textsuperscript{2} &
Yanis Labrak\textsuperscript{6} &
Sameer Khurana\textsuperscript{3,$\dagger$} \\[0.7em]
Antoine Laurent\textsuperscript{5,$\dagger$} &
Mickael Rouvier\textsuperscript{4} &
Phil Woodland\textsuperscript{2} &
Ricard Marxer\textsuperscript{1,7}
\end{tabular}}\\[1.9em]
\makebox[\dimexpr\textwidth-2\tabcolsep\relax][c]{\normalfont\small
\parbox{0.98\textwidth}{\centering
\textsuperscript{1}Universit\'e de Toulon, Aix-Marseille Universit\'e, CNRS, LIS; 
\textsuperscript{2}Department of Engineering, University of Cambridge; 
\textsuperscript{3}Mitsubishi Electric Research Laboratories; 
\textsuperscript{4}LIA, Avignon Universit\'e; 
\textsuperscript{5}LIUM, Le Mans Universit\'e; 
\textsuperscript{6}Idiap Research Institute; 
\textsuperscript{7}CNRS, ILLS\\ 
\textsuperscript{$\dagger$}Contributed while at the listed affiliations.}}
}

\begin{document}

\ifcolmsubmission
\linenumbers
\fi

\maketitle

\begin{abstract}
Text-pretrained language models (LMs) encode rich world knowledge, but adapting them to process and generate perceptual modalities such as audio and images while effectively leveraging that knowledge remains challenging. Perceptual modalities are finer-grained and less semantically dense than text, making it unclear how functions learned during text pretraining can be reused. We study this problem through the lens of a layerwise abstraction-refinement dynamic observed in transformer LMs: representations first become more abstract and compositional, then are refined into representations predictive of fine-grained structure. This perspective suggests that adapting an LM to finer-grained modalities requires: (i) allocating additional fine-to-coarse processing at the input and coarse-to-fine processing at the output, consistent with late modality fusion and an output-side analogue we term late fission; and (ii) allowing the output predictor to preserve input-dependent selective access to both high-level semantic structure and low-level perceptual detail, motivating our use of attention residuals in fission. We instantiate this view in LF\textsuperscript{2}AR, a simple architecture combining these mechanisms, and study it on text-as-images and speech, two modalities for which semantic correspondence to text can be controlled. Across models ranging from 135M to 2B parameters and adapted to these modalities, we find that these components increase feature abstraction, strengthen cross-modal alignment, enable such alignment to emerge at smaller compute budgets, improve preservation of text-like predictive structure, and yield better performance on text-as-images and speech versions of language understanding and reasoning benchmarks. Additionally, attention residuals induce sparse, interpretable use of deep backbone layers, enabling early-exit decoding with a 1.9$\times$ generation speedup.
\end{abstract}

\section{Introduction}

The remarkable capabilities of large language models (LLMs) trained on web-scale text corpora, together with the relative scarcity of equally rich semantic supervision in other modalities, have made transfer of text-pretrained knowledge a central goal in multimodal AI \citep{alayrac2022flamingo,blip2-li-2023,Girdhar_2023_CVPR_imagebind,shi2025lmfusion}. A common strategy is to fine-tune pretrained text LLMs on multimodal corpora and instruction-tuning data so that they can consume and generate non-textual signals while reusing capabilities acquired during text pretraining. This recipe underlies much recent progress in multimodal language modeling. Yet the resulting transfer is often imperfect: models frequently underperform when semantically equivalent content is presented in non-textual rather than textual form \citep{tang2025seam,venhoff2025too,cuervo2026closing}, and multimodal adaptation can also induce forgetting of text-only capabilities \citep{zhang2024wings, cuervo2026closing}. These failures highlight the challenge of adapting LLMs to multimodal signals in a way that preserves and reuses what was learned during text pretraining.

We study the challenge of multimodal adaptation primarily as one of cross-modal transfer: semantically equivalent content should engage similar computations across modalities. For example, the sentence \textit{“Your submission has been accepted”} should light up similar neural circuits of positive emotions, whether heard aloud or read in an email. A central obstacle to such transfer is a mismatch in representational granularity. Text token sequences are semantically dense, with each token typically corresponding to a word-like unit. By contrast, visual or audio signals are represented as much longer streams of lower-level units---pixels or patches and audio frames, respectively---over which semantics are distributed more sparsely along spatial or temporal dimensions and mixed with modality-specific information. As a result, cross-modal transfer is non-trivial: even when the underlying semantics are the same, the representation space of perceptual modalities differs substantially from the input space over which text-learned functions are defined.

A large body of prior work has attempted to address these challenges through complex multi-stage training pipelines and modular architectures with adapters that bridge the granularity mismatch between perceptual signals and language representations, both at the input---a family of methods often termed as doing \emph{late modality fusion}~\citep{alayrac2022flamingo,blip2-li-2023,jiang2024maven,Yang2024VisionZipLI,shi2025lmfusion,tseng2026taste}---and at the output~\citep{blip2-li-2023,dong2024dreamllm, xu2025qwen25omnitechnicalreport}, which we analogously term \emph{late modality fission}\footnote{Perhaps a non-standard term, but with precedents in the literature \citep{landragin:halshs-00138503, Marinov_2023_ICCV}.}. Many other works, however, adopt and advocate for simpler, more monolithic architectures that rely on lightweight linear mappings for multimodal adaptation~\citep{chameleonteam2024chameleonmixedmodalearlyfusionfoundation,nguyen-etal-2025-spirit,zeng2025scaling,janus_Wu_2025_CVPR,cui2025emu35nativemultimodalmodels}. These methods are commonly referred to as \emph{native} multimodal LMs or, in our preferred terminology, as performing \emph{early modality fusion}; correspondingly, we refer to their output-side analogue as \emph{early modality fission}. The trade-offs between these approaches remain an active area of research~\citep{Shukor_2025_ICCV,venhoff2025too,diao2026towardsnativeneo}.

In this work, rather than starting from complex multimodal adaptation pipelines or from the early-versus-late fusion/fission dichotomy alone, we ask what the internal organization of a text-pretrained LM implies about how multimodal adaptation should be structured. Our starting point is recent work suggesting that transformer LMs exhibit a consistent two-phase representational dynamic across layers: earlier layers tend to compose lower-level features into more abstract and compositional ones, whereas later layers refine these abstractions into lower-level features predictive of upcoming finer-grained structure \citep{valeriani2023the,antonello2024evidence,cheng2025emergence,queipo-de-llano2026attention}. We refer to this as the \textit{layerwise abstraction-refinement dynamics}. Because perceptual signals are finer-grained than text, effective adaptation that preserves and leverages this dynamic should provide additional modality-specific layers for fine-to-coarse processing before the pretrained backbone and coarse-to-fine processing after it. We study these as simple instances of \emph{late fusion} and \emph{late fission}, respectively, and analyze their effects on representation dynamics across layers to better understand their relation to cross-modal transfer.

Additionally, we hypothesize that, given the finer granularity of perceptual signals in time and/or space relative to text, generating them should rely only sparsely on higher-level semantic features and more often on low-level local detail. For example, in the context of speech generation, given the utterance \texttt{``we submitted the paper to the ''}, predicting upcoming speech tokens may initially benefit from high-level features, such as anticipating \texttt{``conference''}. However, once the model begins generating \texttt{``conference''}, accurate prediction depends on fine-grained local context---for instance, knowing that the current phone is \texttt{`f`} is necessary to predict the next phone \texttt{`e`}. This leads us to propose a novel design primitive based on \emph{attention residuals}~(similar to \citet{kimiteam2026attentionresiduals}), which gives the fission mechanism dynamic access to multiple representation levels and allows it to switch between coarse semantic information and fine perceptual detail as needed.

Our main contributions can be summarized as follows:

\begin{itemize}
    \item We introduce simple architectural instantiations of late fusion and late fission, together with a framework for studying their effects on cross-modal transfer through the lens of layerwise abstraction-refinement dynamics.
    \item We propose an additional design primitive for multimodal adaptation of LMs: attention residuals to augment modality fission.
    \item We apply this framework to the study of text LMs adapted for generation of text-as-images and speech, and through extensive ablations demonstrate the effects of these multimodal adaptation architectural choices.
    \item We show that fission with attention residuals learns sparse use of backbone depth, and leverage it to speed up generation through dynamic compute allocation.
\end{itemize}

\section{Preliminaries}

\paragraph{Multimodal adaptation of language models.} Let $\mathbf{w} = (w_1,\dots,w_T)$ denote a sequence of text tokens, with $w_i \in \mathcal{V}_{\text{text}}$. A text LM $P_\theta$ defines the autoregressive distribution
$
P_\theta(\mathbf{w}) = \prod_{i=1}^{T} P_\theta(w_i \mid \mathbf{w}_{<i}),
$
and is pretrained by maximum likelihood over a text corpus, $\mathcal{D}_{\text{text}}$:
$
\mathcal{L}_{\text{text}}(\theta)
=
-\sum_{\mathbf{w} \in \mathcal{D}_{\text{text}}}
\sum_{i=1}^{|\mathbf{w}|}
\log P_\theta(w_i \mid \mathbf{w}_{<i})
\label{eq:text_nll}
$. We study \emph{multimodal adaptation} as a second stage of training applied to such a pretrained text LM. Our focus is on models adapted not only for multimodal understanding, but also for multimodal generation, through continued autoregressive modeling of sequences interleaving text with tokens from an additional modality. Let $\mathbf{x} = (x_1,\dots,x_T)$ denote such a sequence, where each token belongs either to the text vocabulary $\mathcal{V}_{\text{text}}$ or to the vocabulary $\mathcal{V}_{m}$ of one target modality $m$, i.e.,
$
x_i \in \mathcal{V}_{\text{mm}} = \mathcal{V}_{\text{text}} \cup \mathcal{V}_{m}.
$
Starting from the pretrained LM, multimodal adaptation yields a model $P_{\theta'}$, where $\theta'$ may include both the original backbone parameters and any additional modality-specific parameters. $P_{\theta'}$ is trained to autoregressively model $\mathbf{x}$:
\begin{equation}
\mathcal{L}_{\text{mm}}(\theta')
=
-\sum_{\mathbf{x} \in \mathcal{D}_{\text{mm}}}
\sum_{i=1}^{|\mathbf{x}|}
\log P_{\theta'}(x_i \mid \mathbf{x}_{<i}).
\label{eq:mm_nll}
\end{equation}

In this paper, we study two concrete instances of this setup: LMs adapted for \emph{pixel language modeling} \citep{rust2023language} and \emph{speech language modeling} \citep{lakhotia-etal-2021-generative}. In both cases, multimodal sequences are constructed by switching between text and the target modality at the word level according to a fixed interleaving rate, similarly to \citet{nguyen-etal-2025-spirit}. In the pixel LM setting, the additional modality consists of rendered text images represented as sequences of flattened binary image patches $\mathbf{p}_j \in \{0,1\}^{K}$. The adapted model predicts each patch through a factorized Bernoulli distribution over its pixels,
$
P_{\theta'}(\mathbf{p}_j \mid \mathbf{x}_{<j}) = \prod_{k=1}^{K} P_{\theta'}(p_{j,k} \mid \mathbf{x}_{<j}),
$
conditioned on the preceding multimodal context. In the speech LM setting, the additional modality consists of vector-quantized self-supervised speech representations, as in \citet{lakhotia-etal-2021-generative}. These speech tokens are treated as elements of a modality-specific vocabulary $\mathcal{V}_{\text{speech}}$.


\paragraph{Abstraction-refinement dynamics in transformer language models.} Recent work suggests a consistent picture of how transformer LMs organize computation across depth. Rather than changing monotonically, hidden representations appear to pass through two broad phases, which we refer to jointly as the \textit{layerwise abstraction-refinement dynamic}. In the \emph{abstraction} phase, lower-level features are progressively composed into coarser and more abstract ones; geometric analyses based on intrinsic dimensionality and neighborhood structure associate this phase with an expansion of the representation manifold, suggesting that the model is constructing a richer feature basis in which linguistic structure can be represented~\citep{valeriani2023the,cheng2025emergence}. A useful intuition comes from the compositional nature of language: in a character-level LM, for example, a vocabulary of a few dozen symbols must be composed into a far larger space of morphemes, words, and multi-word meanings. In the subsequent \emph{refinement} phase, the model increasingly shifts toward deriving lower-level features conditioned on the coarser representations composed earlier, in ways that are useful for next token prediction~\citep{cheng2025emergence,antonello2024evidence,queipo-de-llano2026attention}.

\section{Accounting for Layerwise Abstraction-Refinement Dynamics}
\label{sec:arch_hypothesis}

\paragraph{Architectural hypotheses.}A lesson from the abstraction-refinement literature is that different layers tend to operate at characteristic abstraction levels. Consequently, text-pretrained computations are not uniformly reusable across depth: a circuit trained to respond to word-level features, for example, may fail to activate if the representations reaching it are characters. Because perceptual modalities are finer-grained than text, this view suggests the need for additional processing on both sides of the pretrained backbone.

\begin{hypothesisbox}
\paragraph{Hyp.}
Multimodal adaptation benefits from late fusion for additional fine-to-coarse processing before the pretrained text backbone and late fission for additional coarse-to-fine processing after it to better match expected layerwise abstraction levels.
\end{hypothesisbox}

The abstraction-refinement literature also suggests that such additional composition and refinement need not rely on the specialized training algorithms and architectures often used in late fusion/fission. The standard language modeling objective with transformer layers already appears sufficient to induce these phases in text-pretrained backbones. We therefore adopt such a minimalist approach, implementing late fusion and late fission as additional layers placed before and after the backbone, respectively, applied only to the target modality, and trained end-to-end with the same autoregressive objective as the rest of the model.

Regarding how fine-grained modeling should use the pretrained model, we posit that high-level text features may be most useful when selecting the next linguistic unit, whereas low-level local structure may matter more when unfolding that unit into a sequence of perceptual features. If so, effective adaptation should avoid forcing such low-level detail to be propagated throughout the pretrained backbone, where it may interfere with the formation and reuse of higher-level text-like features. For example, as illustrated in the introduction, in speech generation, predicting the beginning of a word may benefit from text-like features that anticipate the next lexical item, while predicting the following phones within that word depends on local phonetic context.

\begin{hypothesisbox}
\paragraph{Hyp.} Fine-grained modeling should rely on both higher-level guidance and lower-level detail as needed; thus fission benefits from dynamic access to multiple representation levels, rather than propagating low-level detail through the pretrained backbone.
\end{hypothesisbox}

We instantiate this hypothesis through \emph{attention residuals}. Let $\mathbf{c}_i^{(1)}, \dots, \mathbf{c}_i^{(L)} \in \mathbb{R}^d$ denote the residual-stream\footnote{For potentially useful background on the transformer architecture, please refer to Appendix~\ref{app:transformer-background}.} representations at position $i$ from the $L$ layers of the model up to the backbone output. In a standard residual operation, a layer augments its current representation with a fixed skip contribution from an earlier computation. Here, we use the same basic idea, but make the residual input-dependent: rather than adding a fixed skip connection, we let the fission module attend over the residual states of previous layers and form an input-dependent weighted residual summary. In our implementation, we first construct a lightweight summary
$
\mathbf{c}_i' = \sum_{l=1}^{L} \phi^{(l)} \mathbf{c}_i^{(l)},
$
where $\phi^{(l)} \in \mathbb{R}$ are learned fixed scalar weights. A selector $S : \mathbb{R}^d \rightarrow \mathbb{R}^L$ then uses this summary to produce input-dependent layer attention weights, which are used to compute the attention residual $\bar{\mathbf{c}}_i$:
\begin{equation}
\boldsymbol{\omega}_i = \mathrm{Softmax}(S(\mathbf{c}_i')),
\qquad
\bar{\mathbf{c}}_i = \sum_{l=1}^{L} \omega_i^{(l)} \mathbf{c}_i^{(l)},
\qquad
\tilde{\mathbf{c}}_i = \mathbf{c}_i^{(L)} + \bar{\mathbf{c}}_i.
\label{eq:attn_residuals}
\end{equation}

In this way, $\tilde{\mathbf{c}}_i$, used as input to the modality fission process, allows it to draw selectively on whichever abstraction level is most useful at each step.

Figure~\ref{fig:lf2ar-architecture} illustrates LF\textsuperscript{2}AR (Late Fusion and Late Fission with Attention Residuals), the studied architecture, in the speech setting.

\input{img/lf2ar_architecture_inline_tikz}

\paragraph{Analysis tools.}Our architectural hypotheses are framed in terms of \emph{benefit}: late fusion, late fission, and attention residuals should help multimodal adaptation if they better support cross-modal transfer. To make this concrete, we next introduce a set of layerwise metrics that let us study these benefits in the model's internal representations.

\begin{itemize}[itemsep=0pt, topsep=1pt, leftmargin=10pt]
    \item \textbf{Dataset intrinsic dimensionality.}
Following the abstraction-refinement literature, we use the intrinsic dimensionality of the dataset representation manifold as a geometric proxy for feature abstraction and compositionality~\citep{valeriani2023the}. For each layer $l$, let $\mathcal{H}^{(l)} = \{\mathbf{h}_n^{(l)}\}_{n=1}^{N}$ denote the set of sample representations extracted from that layer. We estimate the intrinsic dimensionality of $\mathcal{H}^{(l)}$, using the \textit{Generalized Ratios Intrinsic Dimension Estimator} (GRIDE) \citep{denti2022generalized}. Higher intrinsic dimension is interpreted as evidence of a richer, more compositional feature basis.
    \item \textbf{Word-level cross-modal similarity.}
Because our multimodal sequences are interleaved at the word level, we can use alignments between text tokens and modality tokens corresponding to the same linguistic unit. Let $\mathcal{B}$ denote the set of aligned word boundaries, and let $\mathbf{h}^{(l)}_{\text{text},j}$ and $\mathbf{h}^{(l)}_{\text{mod},j}$ denote the text and target-modality representations at boundary $j \in \mathcal{B}$ in layer $l$. We define word-level cross-modal similarity as
$\frac{1}{|\mathcal{B}|}
\sum_{j \in \mathcal{B}}
\cos\!\left(
\mathbf{h}^{(l)}_{\text{text},j},
\mathbf{h}^{(l)}_{\text{mod},j}
\right)$. This gives a simple measure of cross-modal alignment at the level of linguistic units.
    \item \textbf{BERT alignment.} To measure semanticity, we compare the neighborhood structure of target-modality representations to that induced by a pretrained BERT-style text embedding model. For each layer $l$, we compute the $k$-nearest-neighbor graphs of target-modality representations and aligned text semantic embeddings across samples in $\mathcal{H}^{(l)}$, and compute their overlap. Higher values indicate more semantic representations.
    \item We train TunedLens-style affine probes~\citep{belrose2025elicitinglatentpredictionstransformers} at every layer. We consider three prediction tasks:
    \begin{itemize}
        \item \textbf{Current-modality-token perplexity} measures how easily the current perceptual unit can be reconstructed from the representation. Lower perplexity indicates that more low-level modality-specific information remains available. Because the model's unembedding layer is trained for next-token prediction rather than decoding the current modality token, we use a separate learned output projection.
        \item \textbf{Next-modality-token perplexity} measures how far the representation has been refined toward predicting the next fine-grained target-modality unit.
        \item \textbf{Next-text-token perplexity} is measured from target-modality representations at aligned word boundaries. Lower perplexity indicates stronger preservation of the text-like next-token predictions learned during pretraining.
    \end{itemize}
\end{itemize}

For dataset intrinsic dimensionality, BERT alignment, and next-text-token perplexity, we also report as reference the corresponding values obtained from text inputs.

\section{Experimental Setup}

\paragraph{Models.}We conduct our experiments on SmolLM backbones~\citep{allal2025smollm}, available in three sizes---135M, 360M, and 1.7B---which we adapt to process and generate text-as-images or speech. To study our architectural hypotheses, we train model variants ranging from an early fusion/fission model, denoted \emph{Base}, to a late fusion, late fission model with attention residuals, denoted \emph{LF²AR}. To assess both component-wise effects and their interactions, we use complementary ablations: additive ablations from \emph{Base} for pixel-adapted LMs, and subtractive ablations from \emph{LF²AR} for speech-adapted LMs. We implement late fusion and late fission as two decoder layers each, with the same architecture as the backbone, placed before and after it, respectively. In preliminary experiments, additional layers brought little gain in dataset intrinsic dimensionality, so we kept this configuration throughout. We use the \texttt{MiniLM-L6-v2}\footnote{\url{https://huggingface.co/sentence-transformers/all-MiniLM-L6-v2}} embedding model for the calculation of BERT alignment.

\paragraph{Data.}For pixel-adapted LMs, we train on the English subset of FineWiki~\citep{penedo2025finewiki}, comprising 6.6M documents. We render it as images using the renderer of \citet{lotz-etal-2023-text}, yielding 7.04B image patches, together with word-level alignments. For speech-adapted LMs, we train on a collection of public English speech datasets, described in Appendix~\ref{app:setup}, totaling 160k hours of audio. We tokenize them using the speech tokenizer of \citet{hassid2023textually}, yielding 10.89B speech tokens, and use the forced aligner of \citet{pratap24scaling} to obtain word timestamps. All representation-analysis metrics are reported on a held-out set.

\paragraph{Training.}All models are trained end-to-end to minimize Equation~\ref{eq:mm_nll} using the AdamW optimizer~\citep{loshchilov2018decoupled}. We train the models adapted from SmolLM 135M and 360M for up to 16B tokens. We use the 1.7B backbone only for speech and train it for up to 32B tokens. We adopt Warmup-Stable-Decay learning-rate schedules, with different peak learning rates for the backbone and the late fusion/fission layers. We tuned the peak learning rates separately for each setup, selecting the largest stable values.

\paragraph{Benchmarks.} We evaluate on multiple-choice benchmarks commonly used for LLMs: image-rendered versions of MMLU~\citep{hendrycks2021measuring}, HellaSwag~\citep{zellers2019hellaswag}, and StoryCloze~\citep{mostafazadeh-etal-2016-corpus} for pixel LMs, and the spoken version of StoryCloze~\citep{hassid2023textually} for speech LMs. For all benchmarks, we also implement cross-modal variants in which questions and answers may be presented in different modalities. We report the accuracy with which the model assigns higher likelihood to the correct answer.

See Appendix~\ref{app:setup} for more details on the experimental setup. 

\begin{figure}[t]
\begin{center}
\centerline{\includegraphics[width=\textwidth]{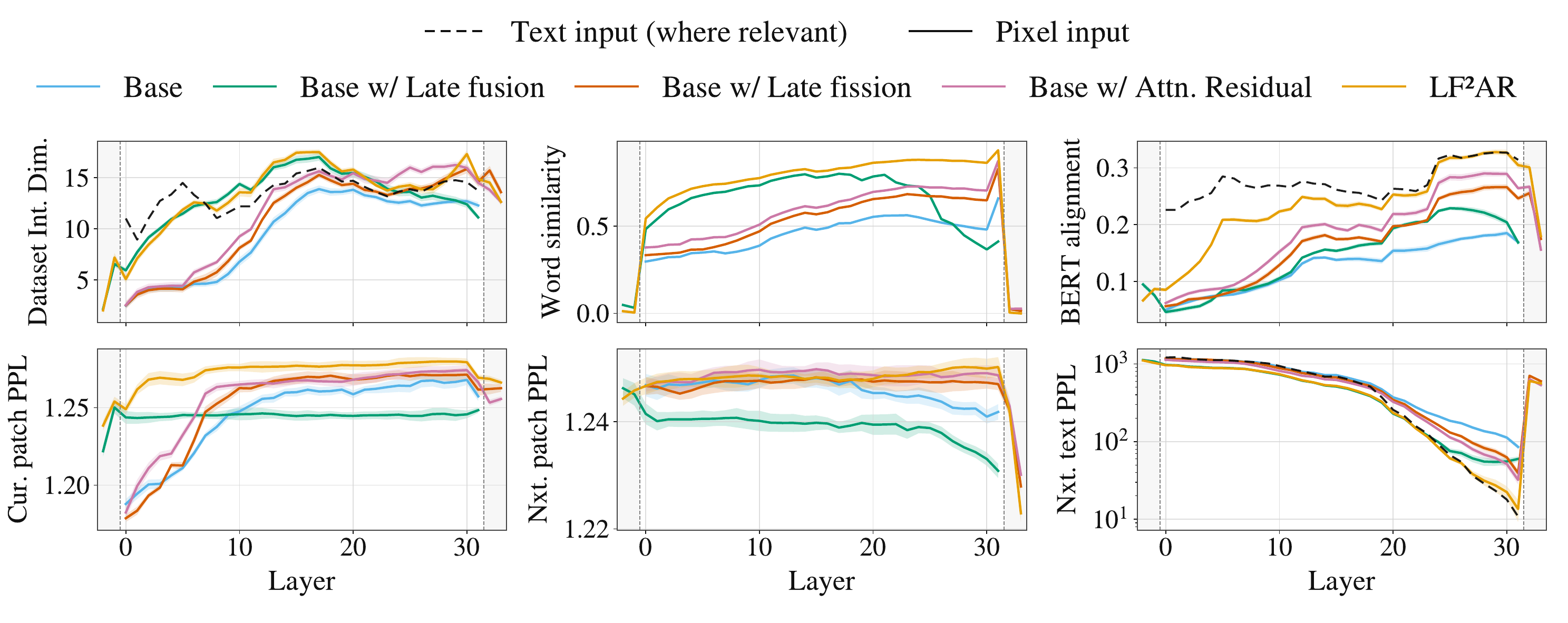}}
\vskip -0.15in
\caption{
    Layerwise analyses for pixel-adapted LMs across additive ablations.
    Late fusion improves early abstraction and alignment, late
    fission preserves deeper text-like predictive structure, and attention
    residuals provide additional gains. Solid curves indicate the means and shaded regions indicate $\pm 1$ standard deviation across
    five 1,024-sample subsets.
}
\label{fig:analysis_pixlm}
\centerline{\includegraphics[width=\textwidth]{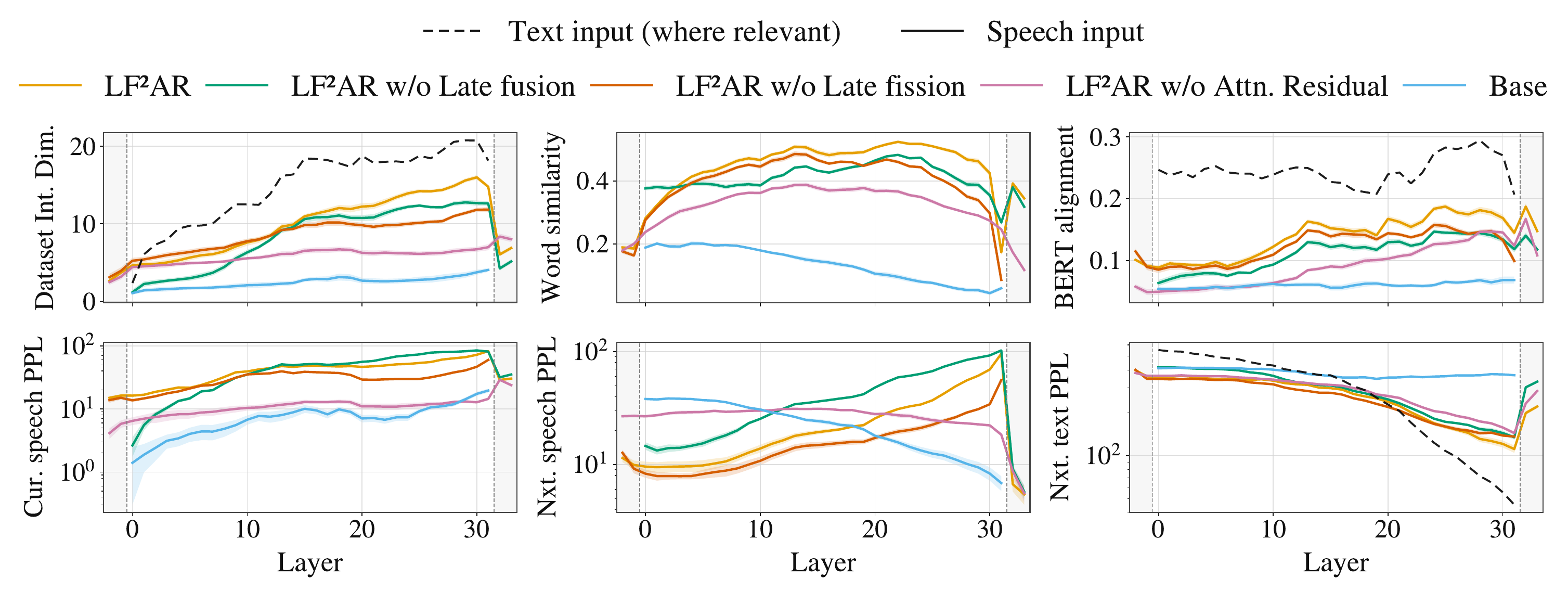}}
\vskip -0.15in
\caption{Layerwise analyses for speech-adapted LMs across subtractive ablations. Removing late fusion/fission or attention residuals lowers abstraction, alignment, and semanticity, with attention residuals having the strongest effect on retaining low-level speech detail. Solid curves indicate the means and shaded regions indicate $\pm 1$ standard deviation across five 1,024-sample subsets.}
\label{fig:analysis_speechlm}
\end{center}
\vskip -0.3in
\end{figure}

\section{Results} 
\label{sec:results}

\paragraph{Representation analyses.} We analyze the layerwise metrics across the different ablated model variants. For this experiment, we use only the 360M backbone. Figures~\ref{fig:analysis_pixlm} and~\ref{fig:analysis_speechlm} present the results for the pixel-adapted and speech-adapted LMs, respectively. In both modalities, \emph{Base} and \emph{LF²AR} score on average the lowest and highest, respectively, in compositionality, as measured by dataset intrinsic dimensionality, word-level cross-modal alignment, and semanticity as measured by BERT alignment. Overall, \emph{Base} also exhibits stronger retention of low-level information and worse language-modeling performance. Notably, \emph{LF²AR} image representations are as semantic and predictive as text representations.

Figure~\ref{fig:analysis_pixlm} shows that, for pixel adaptation, adding late fusion yields higher compositionality and word-level cross-modal alignment. This effect persists through roughly the first two-thirds of the model depth, after which both metrics drop toward \emph{Base} levels. At that point, next-image-patch perplexity indicates that the model begins to specialize in fine-grained prediction. In the deeper layers, late fusion also yields representations with higher semanticity than \emph{Base} and that are more predictive of language, as shown by next-text-token perplexity. The addition of late fusion also leads to a rapid early loss of low-level detail. Figure~\ref{fig:analysis_speechlm} shows a similar picture for speech adaptation: removing late fusion produces early-layer representations that are less abstract and less semantic, while also retaining more low-level detail and exhibiting overall worse language-modeling performance.

\begin{keyinsight}
\paragraph{Key insight:} Late fusion induces more text-like early-layer features, leading to more semantic intermediate representations and improved cross-modal transfer.
\end{keyinsight}

\begin{figure}[t]
\begin{center}
\centerline{\includegraphics[width=\textwidth]{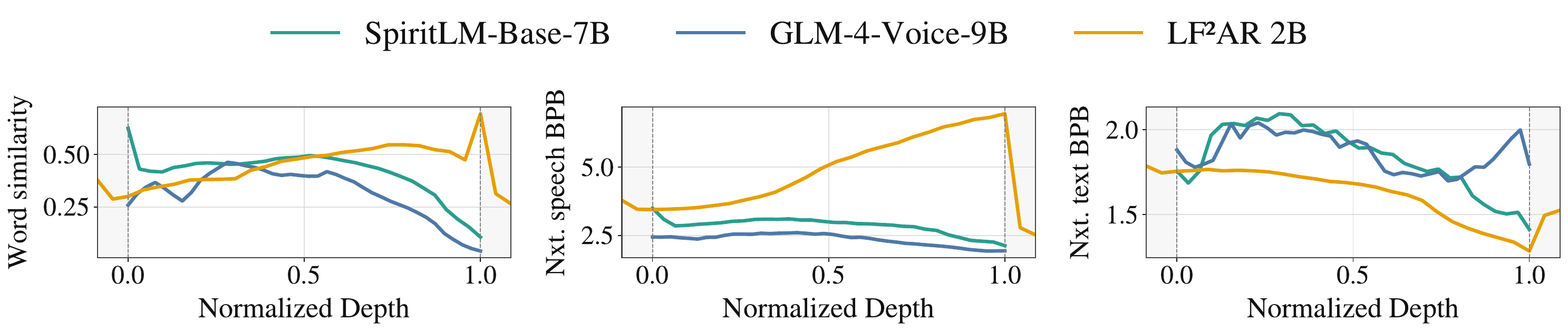}}
\vskip -0.15in
\caption{In open-weight speech LLMs with early fission, deeper layers become specialized for fine-grained speech prediction and lose next-word-predictive structure.}
\label{fig:fission-problem-literature}
\end{center}
\vskip -0.3in
\end{figure}

The clearest effect of late fission is to eliminate the specialization of the deepest layers for fine-grained prediction, which in turn yields higher compositionality, cross-modal alignment, semanticity, and greater use of high-level information predictive of text. In speech, removing late fission produces the corresponding effect, reducing these metrics in the deepest layers. Interestingly, Figure~\ref{fig:fission-problem-literature} illustrates the same phenomenon in the open-weight speech-adapted LLMs SpiritLM-7B~\citep{nguyen-etal-2025-spirit} and GLM-4-Voice-9B~\citep{zeng2025scaling}, showing that specialization of the deepest layers toward the target modality consistently coincides with the loss of next-word-predictive features. Notably, our adaptation of the 1.7B backbone, \emph{LF²AR} 2B, seems\footnote{We qualify this statement because, despite reporting bits per byte, different tokenizations induce factorizations of the sequence distribution of varying prediction difficulty.} to achieve stronger next-text-token prediction despite being a substantially smaller model. We note that, in this comparison, we report bits-per-byte (BPB) rather than perplexity to allow a fairer comparison across models with different tokenizers.

\begin{keyinsight}
\paragraph{Key insight:} Late fission prevents the late layers of the pretrained model from being repurposed for fine-grained modeling, preserving pretrained next-word prediction.
\end{keyinsight}

Adding attention residuals to late fission in pixel-adapted LMs does not alter the layerwise dynamics of late fission alone, but provides a modest consistent boost in compositionality, cross-modal alignment, and semanticity. In speech, however, removing attention residuals leads to a marked increase in the retention of low-level detail and a corresponding drop in compositionality, cross-modal alignment, and semanticity. Across all subtractive ablations, removing attention residuals has the strongest effect on the retention of low-level detail. We hypothesize that this stronger effect in speech reflects its higher entropy relative to artificially rendered text, which causes greater interference with the backbone’s normal function when propagated through it.

\begin{keyinsight}
\paragraph{Key insight:} Fission with attention residuals removes the need to propagate low-level information throughout the pretrained backbone, thereby allowing higher-level features to form and enabling cross-modal transfer.
\end{keyinsight}

For additional experiments on the importance of capacity placement and the dynamic weighting implemented by attention residuals, see Appendices~\ref{app:placement-controls} and~\ref{app:static-residual}.

\begin{figure}[t]
\begin{center}
\centerline{\includegraphics[width=\textwidth]{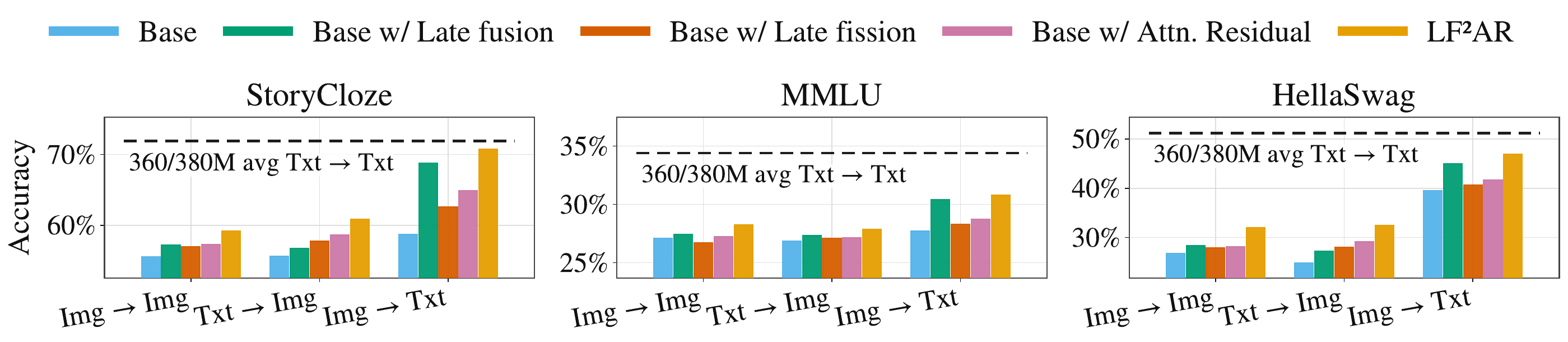}}
\vskip -0.15in
\caption{Downstream benchmark accuracy for pixel-adapted LMs across additive ablations from \emph{Base}. Gains are largest when reasoning must be recovered from image input and expressed in text (\texttt{Img$\rightarrow$Txt}), with late fusion contributing most of the improvement and late fission plus attention residuals adding consistent further gains.}
\label{fig:perf_pixlm_ablation}
\end{center}
\vskip -0.25in
\end{figure}

\paragraph{Effects on downstream performance.} Figures~\ref{fig:perf_pixlm_ablation} and~\ref{fig:speech_pixlm_ablation} show that the above representation-level changes translate into consistent downstream gains. For pixel-adapted LMs, improvements are modest when both input and output remain in the image modality, but much larger when the model must answer in text. Across StoryCloze, MMLU, and HellaSwag, \emph{LF²AR} performs best in the \texttt{Img$\rightarrow$Txt} setting, with late fusion accounting for most of the gain and late fission plus attention residuals adding further improvements. This matches the layerwise analyses: stronger early abstraction and preserved late text-like refinement are especially useful when textual reasoning must be recovered from perceptual input. The pattern holds for speech LMs. \emph{LF²AR} outperforms \emph{Base}, while removing late fusion/fission or attention residuals hurts performance. Late fusion has the largest effect, while late fission and attention residuals provide additional gains, especially in cross-modal settings.

\begin{keyinsight}
\paragraph{Key insight:} Late fusion delivers the largest gains in perceptual understanding, which also improves generation, while late fission is most important for perceptual generation; attention residuals add further gains, especially in cross-modal settings.
\end{keyinsight}

\begin{figure}[t]
\begin{center}
\centerline{\includegraphics[width=\textwidth]{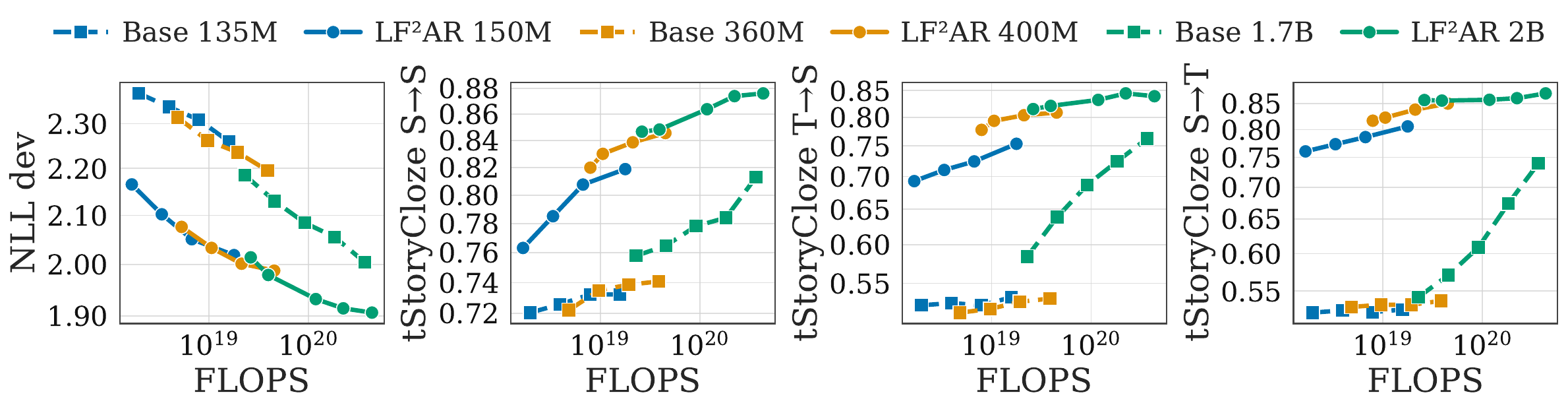}}
\vskip -0.15in
\caption{Scaling curves for speech-adapted LMs. \emph{LF²AR} improves test negative-log-likelihood (NLL) and spoken StoryCloze performance across compute (in FLOPs) scales.}
\label{fig:scaling_perf_speech}
\end{center}
\vskip -0.2in
\end{figure}

Figures~\ref{fig:scaling_perf_speech} and~\ref{fig:scaling_perf_pixlm} show that these gains persist across compute scale. For pixels, \emph{LF²AR} improves over \emph{Base} at both 150M and 375M parameters, with the largest gains on \texttt{Img$\rightarrow$Txt}. For speech adaptation, \emph{LF²AR} scales much more favorably. Most notably, the cross-modal \texttt{T$\rightarrow$S} and \texttt{S$\rightarrow$T} settings improve sharply with scale for \emph{LF²AR}, whereas \emph{Base} remains nearly flat. These architectural gains often exceed what is obtained by scaling the baseline alone: e.g., 150M \emph{LF²AR} models are competitive with or stronger than 360M \emph{Base} models, and 400M \emph{LF²AR} models consistently outperform 1.7B-scale \emph{Base} in cross-modal transfer.

\begin{keyinsight}
\paragraph{Key insight:} Using late fusion and fission with attention residuals often outperforms parameter scaling alone, and makes cross-modal capabilities appear at smaller scale.
\end{keyinsight}

Table~\ref{tab:benchmark} shows that speech-adapted \emph{LF²AR} models achieve strong performance on StoryCloze, competitive with literature baselines trained with orders of magnitude more compute.

\begin{wraptable}[13]{r}{0.4\textwidth}
\vspace{-1em}
  \centering
  \setlength{\tabcolsep}{3pt} 
  \resizebox{0.4\textwidth}{!}{%
    \begin{tabular}{@{}lccccc@{}}
        \toprule
        \multirow{2}{*}{Model} & \multicolumn{4}{c}{t-StoryCloze} & \multirow{2}{*}{Tok/s} \\
        \cmidrule(lr){2-5}
         & \texttt{T} & \texttt{S} & \texttt{T$\to$S} & \texttt{S$\to$T} & \\
        \midrule
        \textsc{LF²AR}+ & 91.0 & \textbf{83.9} & \textbf{79.9} & 84.6 & 25.1 \\
        \makecell[l]{\textsc{LF²AR}+\\w/ early exit} & 91.0 & 83.3 & 79.3 & \textbf{85.0} & \textbf{47.7} \\
        \bottomrule
    \end{tabular}
}
\caption{Early-exit decoding for \emph{LF²AR}. Skipping deeper layers guided by the attention residual weights preserves performance while increasing throughput $1.9\times$.}
\label{tab:early_exit}
\end{wraptable}

\begin{figure}[t]
\begin{center}
\centerline{\includegraphics[width=\textwidth]{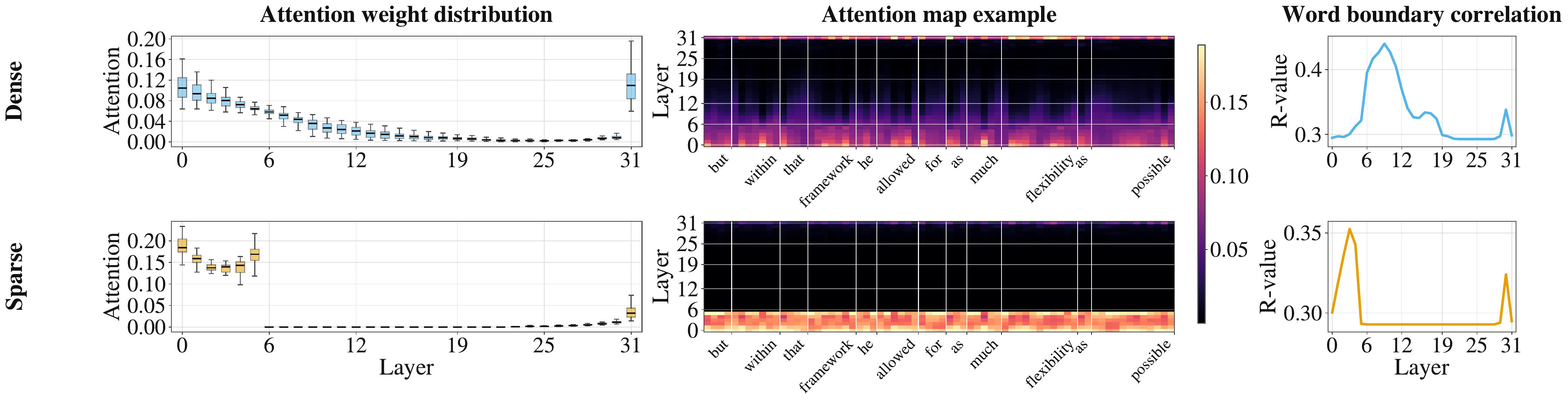}}
\vskip -0.15in
\caption{Attention residuals learn sparse, interpretable backbone depth access. \textbf{Left:} Layer-attention distributions before/after sparsity tuning. \textbf{Middle:} example token-level attention maps. \textbf{Right:} word-boundary detection score using attention spikes as predictors.}
\label{fig:attnres_analysis}
\end{center}
\vskip -0.3in
\end{figure}

\paragraph{Dynamical depth with attention residuals.} Figure~\ref{fig:attnres_analysis} shows that attention residuals learn the two-mode pattern hypothesized in Section~\ref{sec:arch_hypothesis}. Most attention mass concentrates on lower layers and the deepest backbone layer, while intermediate layers receive little attention. The attention maps also show that access to the deepest layer is sparse in the sequence dimension. To interpret this sparsity, we measure the correlation between attention spikes and word boundaries by treating attention peaks as boundary predictions and computing an $R$-value segmentation score~\citep{rasanen2009improved}. As shown in the right panel of Figure~\ref{fig:attnres_analysis}, the layers that receive the most attention are also those whose spikes correlate most strongly with word boundaries. This suggests that attention residuals learn to consult high-level textual features sparsely and at a dynamic rate that matches the changes in linguistic units.

This phenomenon suggests that much of the deep-backbone computation could be made conditional. To test this, we continue training for roughly 100M tokens, retuning the selector to use a fixed weighted summary of only the first six backbone layers and adding auxiliary losses that encourage sparse depth use, yielding the sparser attention patterns in Figure~\ref{fig:attnres_analysis} (bottom). We then train a lightweight router on the same prefix summary to decide whether each token should use a prefix-only approximation or continue through deeper layers. Further details are in Appendix~\ref{app:attnres_extra}. Table~\ref{tab:early_exit} shows that this preserves performance while increasing throughput by $1.9\times$ under autoregressive decoding without KV caching.

\begin{keyinsight}
\paragraph{Key insight:} Modality fission with attention residuals induces sparse backbone usage, enabling variable depth allocation at inference with minimal performance loss.
\end{keyinsight}

\section{Limitations}
\label{sec:limitations}

The scope of this paper is restricted in three respects. First, we study perceptual signals---spoken language and rendered text---for which semantic correspondence to text can be controlled closely enough to support word-level alignment and straightforward cross-modal analyses. Whether the same layer dynamics and architectural choices extend to modalities without such near one-to-one correspondence to text, such as natural visual scenes, remains to be established. Second, we study adaptation of a pretrained text LM, whose layerwise organization is established before exposure to another modality. Native multimodal pretraining may instead organize representations and computations across depth differently from the outset. Third, our experiments are limited to models of up to 2B parameters, and extrapolation to substantially larger scales remains uncertain.

Finally, the throughput gains from dynamic depth allocation do not translate directly to deployment in standard autoregressive inference pipelines. Such pipelines assume a uniform computation path and maintain a key--value cache at every layer for each preceding token. 
Realizing robust serving gains will therefore require specialized inference implementations, as in other conditional-depth methods~\citep{raposo2024mixturedepths}.

\section{Conclusions}

We argued that multimodal adaptation should account for the layerwise abstraction-refinement dynamics of text language models. From this view, we derived simple architectural hypotheses: finer-grained modalities benefit from extra fine-to-coarse processing before the backbone, extra coarse-to-fine processing after it, and dynamic access to multiple backbone representation levels during generation. Across text-as-images and speech, these design choices consistently improved compositionality, cross-modal alignment, preservation of text-like predictive structure, and downstream accuracy, often with gains that exceeded baseline parameter scaling alone. Attention residuals further induced sparse, interpretable use of deep layers and enabled early-exit decoding with a 1.9$\times$ speedup. Taken together, these results show that effective multimodal adaptation benefits from matching the layerwise organization learned during text pretraining.

\section*{Acknowledgments}

Santiago and Ricard are grateful to the French National Research Agency for its support through grant ANR-20-CE23-0012-01 (MIM), as well as to the Institute of Convergence ILCB, supported by grants from France 2030 (ANR-16-CONV-0002) and the Excellence Initiative of Aix-Marseille University (A*MIDEX). Adel is funded by the Cambridge Trust and is grateful to his former supervisor in Avignon, Professor Yannick Estève, for involving him in this project and providing sponsorship. Other authors also received funding from the European Union's Horizon 2020 research and innovation programme under Marie Skłodowska-Curie grant agreement No.~101007666. This work was supported by HPC GENCI-IDRIS under grants AD011014044R1, AD011014044R2, A0161014876, AD011015344, A0161015099, AD011013061R3, and A0161014871.

Finally, we thank the JSALT 2024 workshop and the Center for Language and Speech Processing at Johns Hopkins University for generously hosting the event, where some of the earliest ideas behind this work began to take shape and where fruitful collaborations and friendships emerged in a stimulating and welcoming environment.

\bibliography{references}
\bibliographystyle{colm2026_conference}

\appendix

\section{Related Work}
\label{app:related-work}

Many prior works adopt late-fusion architectures to improve modality adaptation. In the speech domain, relevant examples include \citet{yu24speechencoderllm, chu2023qwenaudioadvancinguniversalaudio, tang2024salmonn}. More recently, several models have focused on decoupling speech generation from the text LM backbone through additional speech-specific output modules~\citep{xu2025qwen25omnitechnicalreport} or modality-specific forward passes~\citep{xie2024miniomnilanguagemodelshear}, an idea aligned with the intuition behind our late-fission approach: non-textual modalities may also benefit from additional output-side processing. Unlike these works, we adopt a much simpler approach to modality fusion and fission and motivate it through the abstraction-refinement phenomenon observed in transformer models, as described in Section~\ref{sec:arch_hypothesis}. We also do not rely on paired text conditioning for speech generation. 

Perhaps most relevant is the work of~\citet{zhao2026towards}, which studies a modality-based layer-splitting architecture to improve the coupling of speech with pretrained text LLMs and similarly reports representation-level and downstream benefits, albeit using a less comprehensive set of measures. Our attention residuals can be viewed as enabling an input-dependent, dynamic form of layer selection rather than relying on a fixed layer-splitting strategy. We further show that this dynamic selection enables variable-depth allocation and can yield decoding speedups. More broadly, relative to prior work, our representation analyses provide additional insight into the roles and effects of different fusion and fission strategies in multimodal adaptation.

\section{Additional Setup Details}
\label{app:setup}

\paragraph{Training.} All models use a weight decay of 0.1. We use a peak learning rate of 3e-4 for backbone parameters in the 135M and 360M models, and 1e-4 for the 1.7B model. The learning rate for non-backbone parameters is uniformly set to 10$\times$ that of the backbone parameters. We use a linear warmup of 100 steps and linearly decay the learning rate over the final 20\% of training. We use a batch size of 1M tokens with a context length of 2048 tokens. Each batch contains equal proportions of target-modality-only, text-only, and interleaved text-and-target-modality sequences. To produce interleaved cross-modal sequences, we segment each utterance into subsequences and interleave spans of random length at runtime: $10$–$30$ words for text and pixel segments and $5$–$15$ words for speech segments. For the affine probes, we use a learning rate of 1.0 and linearly decay it to zero over 1000 training steps, with a batch size of 256 samples.

\paragraph{Data.} For speech-adapted LMs, we use LibriSpeech \citep{librispeech}, LibriLight \citep{librilight}, SWC \citep{swc}, Tedlium \citep{tedlium}, People \citep{people}, Vox Populi \citep{voxpopuli}, and sTinyStories \citep{cuervo-marxer-2024-scaling}. As noted above, we also train the models on text-only data. For this, we use a 12-billion-token subset of the SmolLM corpus \citep{benallal2024smollmcorpus}, with a data distribution matching that used to pretrain the SmolLM models\footnote{As reported in {\small\url{https://github.com/huggingface/smollm/blob/main/pre-training/}}}.

\paragraph{Representation analysis.} We compute all metrics over five subsets of 1,024 samples, independently sampled with replacement, with each sample consisting of 20 words in the corresponding modality. For pixel models, we use a held-out set from FineWiki, amounting to 0.1\% of the training set. For speech, we use data from the StoryCloze benchmark. Curves report the mean across the five subsets, with shaded regions indicating $\pm 1$ standard deviation. For probe-based perplexity metrics, we additionally train each probe with three random seeds and compute the mean and standard deviation jointly across sample subsets and probe seeds. To compute intrinsic dimensionality and BERT alignment, we pool one representation per sample: for intrinsic dimensionality, we take the final representation in the sequence; for BERT alignment, we use average pooling. We use average pooling for BERT alignment because we found that it yields more discriminative results. Since transformer representations exhibit large activation outliers, when computing intrinsic dimensionality and BERT alignment we truncate feature elements (i.e., individual activations) that exceed the 99th percentile across the dataset, following~\citet{huh24platonic}. To estimate intrinsic dimensionality, we use the GRIDE estimator~\citep{denti2022generalized} implemented in \texttt{dadapy}~\citep{glielmo2022dadapy}, following the procedure described by~\citet{cheng2025emergence}.

\paragraph{Benchmarking.} We synthesize target-modality versions of the benchmark text data using the Kokoro-TTS model\footnote{\url{https://github.com/hexgrad/kokoro-82M}} with the \texttt{af-heart} voice, and the same text renderer used to generate the text-as-images training data. All benchmarks are multiple choice. For each task, the model estimates the normalized log probability of each answer given the context, and accuracy is computed by checking whether the highest-scoring option matches the gold answer. Table~\ref{tab:task_prompts} presents the prompt templates used for each task.

\definecolor{lightblue}{RGB}{220,240,255}
\definecolor{lightorange}{RGB}{255,235,210}
\newcommand{\textcol}[1]{\colorbox{lightblue}{#1}}
\newcommand{\anscol}[1]{\colorbox{lightorange}{#1}}

\begin{table}[h]
  \label{tab:task_prompts}
  \centering
  \begin{tabularx}{\textwidth}{@{}X@{}}
    \toprule
    
    \textbf{StoryCloze }\\
    \textcol{\texttt{<story prefix>}} \anscol{\texttt{<answer>}}\\
    \midrule

    \textbf{MMLU}\\
    The following are multiple choice questions (with answers) about \textcol{\texttt{<topic>}}.\\
    Question: \textcol{\texttt{<question>}}\\
    Answer: \anscol{\texttt{<answer>}}\\
    \midrule

    \textbf{HellaSwag }\\
    \textcol{\texttt{<context>}} \anscol{\texttt{<answer>}}\\
    \midrule
    \bottomrule
  \end{tabularx}
  \caption{Evaluation task prompts. Placeholders are shown in \texttt{monospace}. Blue highlights denote text/target-modality inputs, and orange highlights denote continuations where likelihood is evaluated.}
\end{table}
\clearpage

\section{Sparse Selector Tuning and Early Exit}
\label{app:attnres_extra}

\begin{figure}[t]
\begin{center}
\centerline{\includegraphics[width=\textwidth]{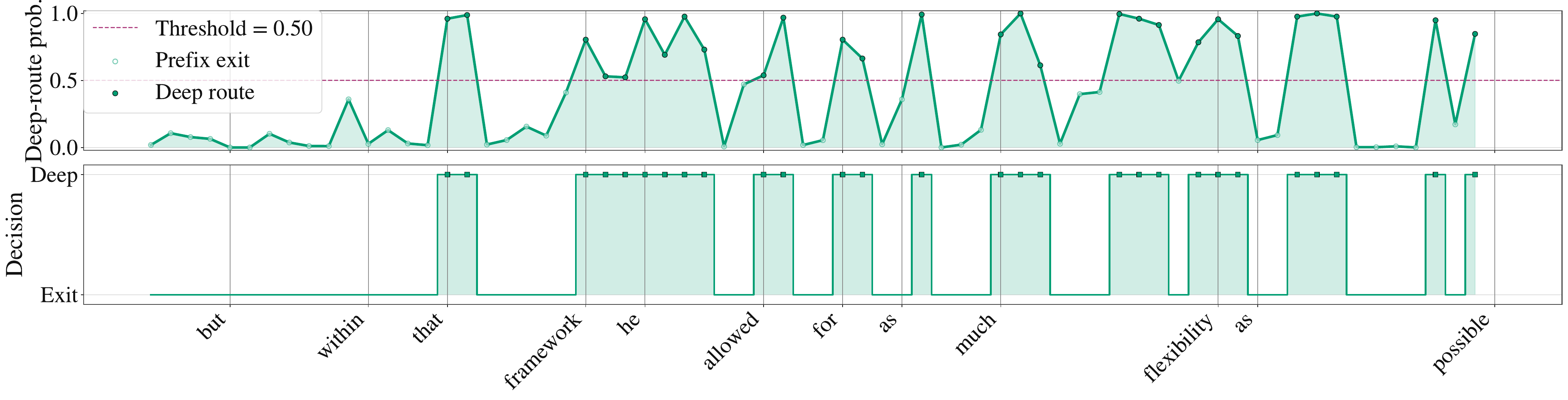}}
\caption{Example router decisions for the sparsity-tuned 360M speech \emph{LF²AR} model. The router, conditioned on a six-layer prefix summary, assigns each token a probability of continuing through the deeper backbone layers (top) and yields a corresponding hard early-exit versus deep-route decision (bottom). Deep routing is used only sparsely along the sequence, illustrating how the model allocates additional compute selectively.}
\label{fig:attnres_router}
\end{center}
\vskip -0.2in
\end{figure}

For the early-exit experiment in Section~\ref{sec:results}, we start from the 360M speech \emph{LF²AR} model and continue training for roughly 100M speech tokens using speech-only batches. During this stage, we freeze the pretrained backbone and update only the speech-specific input/output layers, the fixed layer weights $\phi^{(l)}$, the selector $S$, and, in the routing stage, a small linear router head. 

To make sparse depth decisions available before the full backbone has run, we replace the selector controller input in Equation~\ref{eq:attn_residuals} by a fixed weighted summary of only the first six backbone layers. Concretely, if $\mathbf{c}_i^{(1)},\dots,\mathbf{c}_i^{(6)}$ are the first six residual-stream states, we define
\[
\mathbf{c}'_i=\sum_{l=1}^{6} \tilde{\phi}^{(l)} \mathbf{c}_i^{(l)},
\]
where $\tilde{\phi}^{(l)}$ are the learned fixed layer weights restricted and renormalized to those six layers.

We then add a sparse-selector penalty to the language-modeling objective ($\mathcal{L}_{\mathrm{mm}}$):
\[
\mathcal{L}
=
\mathcal{L}_{\mathrm{mm}}
+
\lambda_D \sum_{l=1}^{L} \frac{l-1}{L-1}\,\omega_i^{(l)}
+
\lambda_{\mathrm{KL}} \mathrm{KL}(\boldsymbol{\omega}_i \,\|\, \mathbf{q}),
\]
averaged over speech positions. We use $\lambda_D=0.05$, and $\lambda_{\mathrm{KL}}=0.02$, with a linear warmup over the first 20 optimization steps. The target distribution $\mathbf{q}$ assigns $0.8$ total mass to the first six layers in proportion to the renormalized fixed layer weights, $0.2$ mass to the final backbone layer, and negligible mass elsewhere.

To obtain actual conditional computation at inference time, we train a lightweight router $r(\mathbf{c}'_i)$ with a binary cross-entropy loss of weight $1.0$. The router target is $1$ iff, in a frozen dense teacher model, the selector peaks on the final backbone layer at position $i$. At inference time, the first six backbone layers are always evaluated. If $\sigma(r(\mathbf{c}'_i)) < 0.5$, the model uses a prefix-only proxy for the audio context; otherwise it continues through the deeper backbone layers. Figure~\ref{fig:attnres_router} illustrates the router predictions for a given sample.

\section{Matched-Capacity Layer-Placement Controls}
\label{app:placement-controls}

A potential alternative explanation for the gains of LF$^2$AR is that they
arise from its additional modality-specific capacity rather than from where that capacity is placed. We therefore conduct a matched-capacity placement ablation in the 400M speech setup. All variants use the same pretrained 360M backbone, the same four newly initialized modality-specific decoder layers, the same attention-residual mechanism, and the same training budget. They differ only in the placement of the additional layers: all four layers before the backbone (2$\times$ input-side), all four layers after the backbone (2$\times$ output-side), or two layers at each interface as in LF$^2$AR.

\begin{figure}[t]
    \centering
    \includegraphics[width=\textwidth]
        {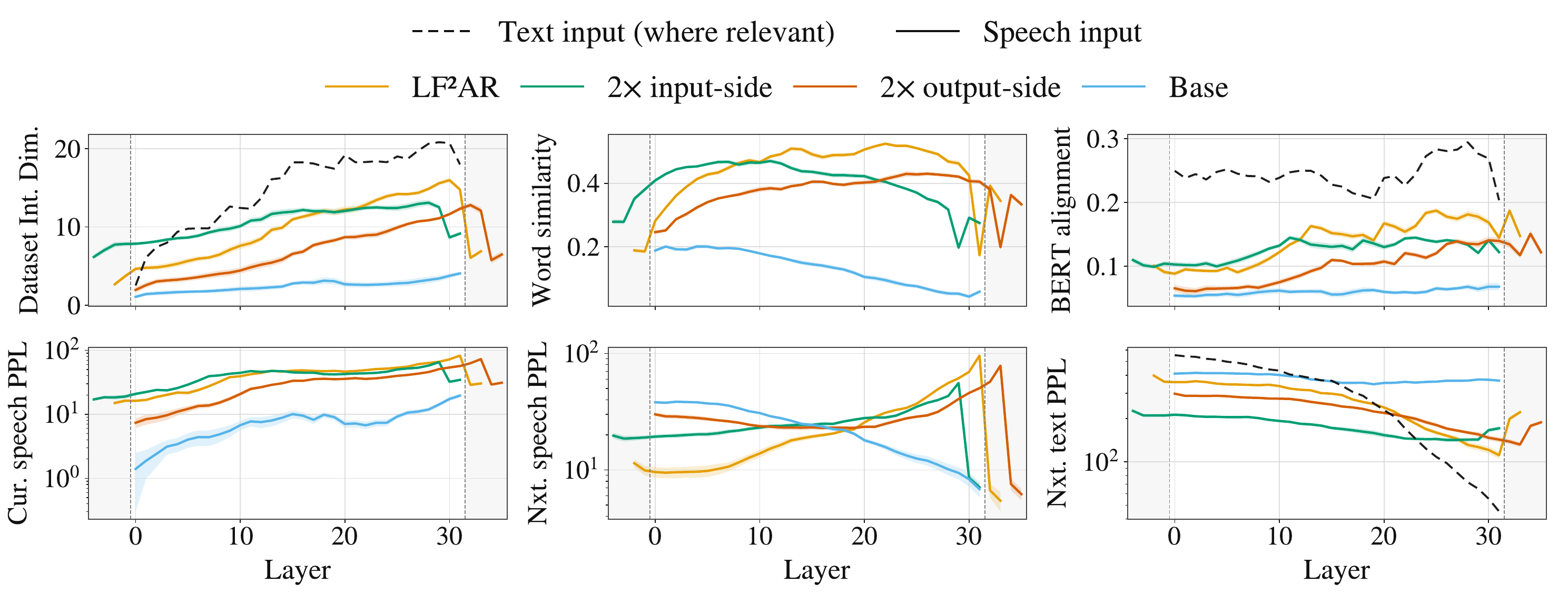}
    \caption{
        Layerwise analyses for matched-capacity placement controls. Input-side placement primarily improves early abstraction and word-level alignment, whereas output-side placement better preserves modality- and text-predictive structure in later layers. Splitting the additional depth across both interfaces, as in LF$^2$AR, yields the strongest overall abstraction, cross-modal alignment, semantic alignment, and reuse of text-learned predictive structure. Solid curves indicate the means and shaded regions indicate $\pm 1$ standard deviation across five 1,024-sample subsets.
    }
    \label{fig:matched-capacity-placement}
\end{figure}

Figure~\ref{fig:matched-capacity-placement} shows that the two controls produce
the directional effects expected from their placement. Concentrating the
additional computation before the backbone produces stronger early
abstraction and word-level alignment than concentrating it after the
backbone. Conversely, the output-side control provides better next-speech and
next-text prediction, consistent with allocating more modality-specific
computation to refinement. Neither placement alone reproduces the complete
LF$^2$AR profile. Splitting the layers across the two interfaces combines
stronger fine-to-coarse processing before the backbone with improved
coarse-to-fine refinement after it.

\begin{table}[t]
    \centering
    \small
    \setlength{\tabcolsep}{6pt}
    \begin{tabular}{lcccc}
        \toprule
        Model
        & \texttt{T} $\uparrow$
        & \texttt{S} $\uparrow$
        & \texttt{T$\rightarrow$S} $\uparrow$
        & \texttt{S$\rightarrow$T} $\uparrow$ \\
        \midrule
        2$\times$ input-side
        & 90.9 & 82.4 & 76.8 & 81.2 \\
        2$\times$ output-side
        & 90.6 & 80.7 & 70.6 & 82.8 \\
        LF$^2$AR
        & \textbf{91.3}
        & \textbf{84.6}
        & \textbf{80.9}
        & \textbf{85.0} \\
        \midrule
        $p$-value vs.\ best placement control
        & 0.54
        & 0.010
        & $1.4{\times}10^{-5}$
        & 0.0096 \\
        \bottomrule
    \end{tabular}
    \caption{
        Downstream performance of the matched-capacity placement
        controls on tStoryCloze.
        All variants have the same newly initialized modality-specific
        capacity. We report two-sided two-proportion tests comparing
        LF$^2$AR with the strongest placement control in each evaluation
        setting.
    }
    \label{tab:matched-capacity-placement-tsc}
\end{table}

As shown in Table~\ref{tab:matched-capacity-placement-tsc}, LF$^2$AR performs
best in all four modality configurations. The difference is small when both
input and output are text, where the pretrained backbone already operates in
its native modality, but is larger in the speech and cross-modal settings.
In particular, the improvements in \texttt{S},
\texttt{T$\rightarrow$S}, and \texttt{S$\rightarrow$T} are significant
relative to the strongest matched-capacity placement control. These results
indicate that the gains are not explained solely by the number of additional
modality-specific layers: where the added computation is placed materially
affects cross-modal transfer.

\section{Fixed Summary Weights and a Static-Residual Alternative}
\label{app:static-residual}

\begin{figure}[t]
    \centering
    \includegraphics[width=0.6\textwidth]
        {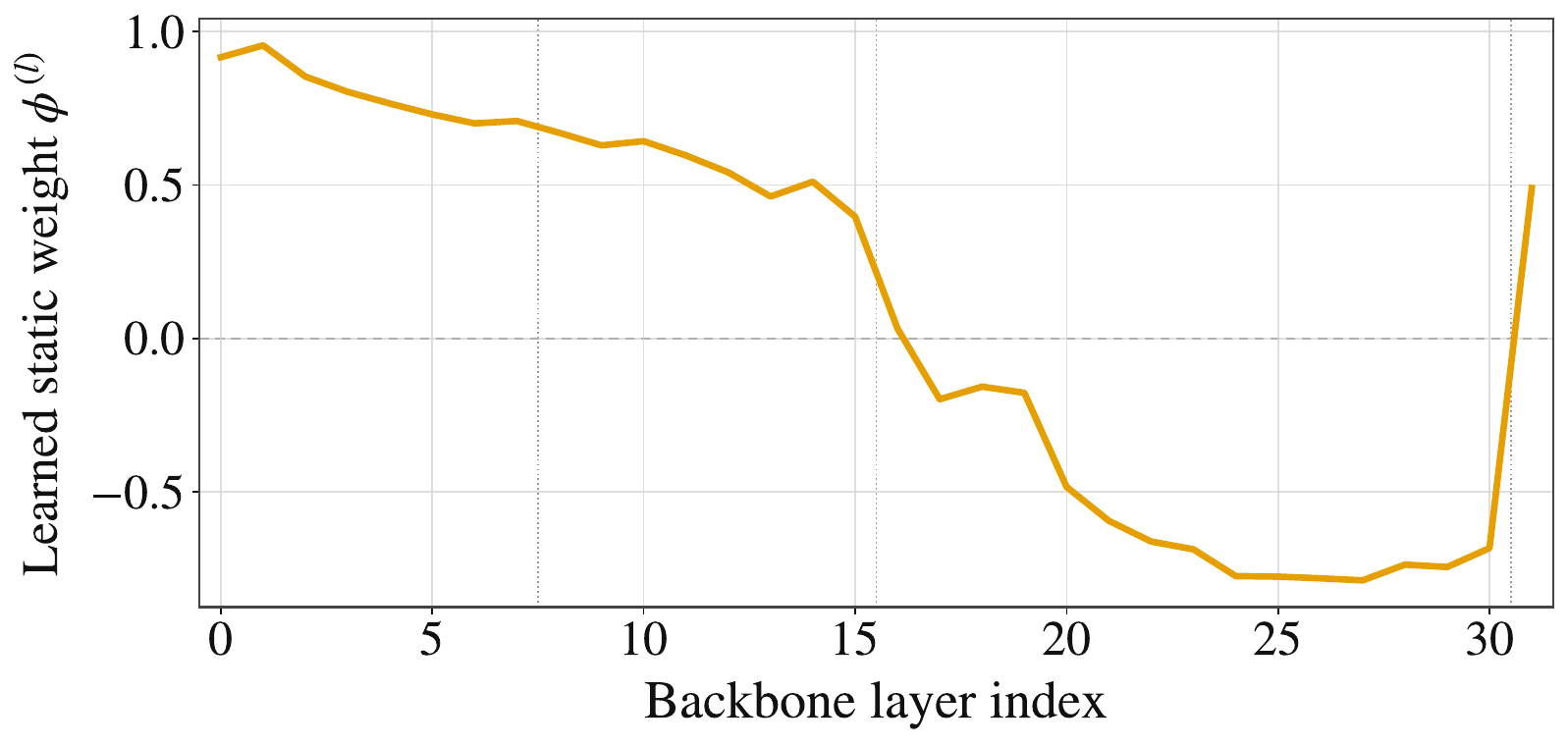}
    \caption{
        Learned fixed summary weights in the main 400M LF$^2$AR
        model. The weights $\phi^{(l)}$ place large positive weight on early
        backbone states, suppress intermediate and deep layers,
        and return to a positive value at the final layer. These are the
        fixed weights used to construct the input to the attention-residual
        selector, rather than the input-dependent selector weights
        $\omega_i$ themselves.
    }
    \label{fig:lf2ar-static-weights}
\end{figure}

The attention-residual mechanism first constructs the lightweight summary
\begin{equation}
    \mathbf{c}'_i
    =
    \sum_{l=1}^{L}
    \phi^{(l)} \mathbf{c}^{(l)}_i,
    \label{eq:app-static-summary}
\end{equation}
where the learned scalars $\phi^{(l)}$ are fixed across inputs. A selector
then uses $\mathbf{c}'_i$ to produce the input-dependent weights
$\omega_i$ used by the fission module. Figure~\ref{fig:lf2ar-static-weights}
shows the learned fixed summary weights from the 400M LF$^2$AR model.

The learned profile exhibits the two-mode organization expected from the
abstraction--refinement view. Early states
provide the selector with low-level and local perceptual information, while the positive final-layer weight provides access to the backbone's most text-predictive representation. Intermediate and deep layers are comparatively suppressed.

To determine whether fixed access to multiple representation levels is
sufficient, we additionally train a static-residual control. This model uses the same pretrained backbone, modality-specific capacity, and training budget as LF$^2$AR, but removes the input-dependent selector and supplies the summary in Equation~\ref{eq:app-static-summary} as the residual to the fission module. The comparison therefore isolates the contribution of dynamically selecting representation depth from the more general benefit of accessing intermediate backbone states.

\begin{table}[t]
    \centering
    \small
    \setlength{\tabcolsep}{6pt}
    \begin{tabular}{lcccc}
        \toprule
        Model
        & \texttt{T} $\uparrow$
        & \texttt{S} $\uparrow$
        & \texttt{T$\rightarrow$S} $\uparrow$
        & \texttt{S$\rightarrow$T} $\uparrow$ \\
        \midrule
        LF$^2$AR-400M
        & \textbf{91.3}
        & \textbf{84.6}
        & \textbf{80.9}
        & \textbf{85.0} \\
        Static residual-400M
        & 90.8
        & 84.0
        & 80.1
        & 83.9 \\
        \bottomrule
    \end{tabular}
    \caption{
        Dynamic versus static residual access on tStoryCloze.
        The static-residual control has access to the same intermediate
        backbone states but cannot select among them as a function of the
        input. Dynamic attention residuals perform best in every modality
        configuration.
    }
    \label{tab:static-residual-tsc}
\end{table}

Table~\ref{tab:static-residual-tsc} shows that fixed residual access recovers part of the benefit but remains consistently weaker than input-dependent attention residuals. The largest difference occurs in
\texttt{S$\rightarrow$T}, where successful transfer requires recovering
text-like predictive structure from speech representations. Moreover, unlike
dynamic attention residuals, the static alternative cannot support
input-dependent depth selection and therefore does not provide the
variable-compute early-exit mechanism.

\clearpage

\section{Additional Figures and Tables}

\begin{figure}[h]
\begin{center}
\centerline{\includegraphics[width=\textwidth]{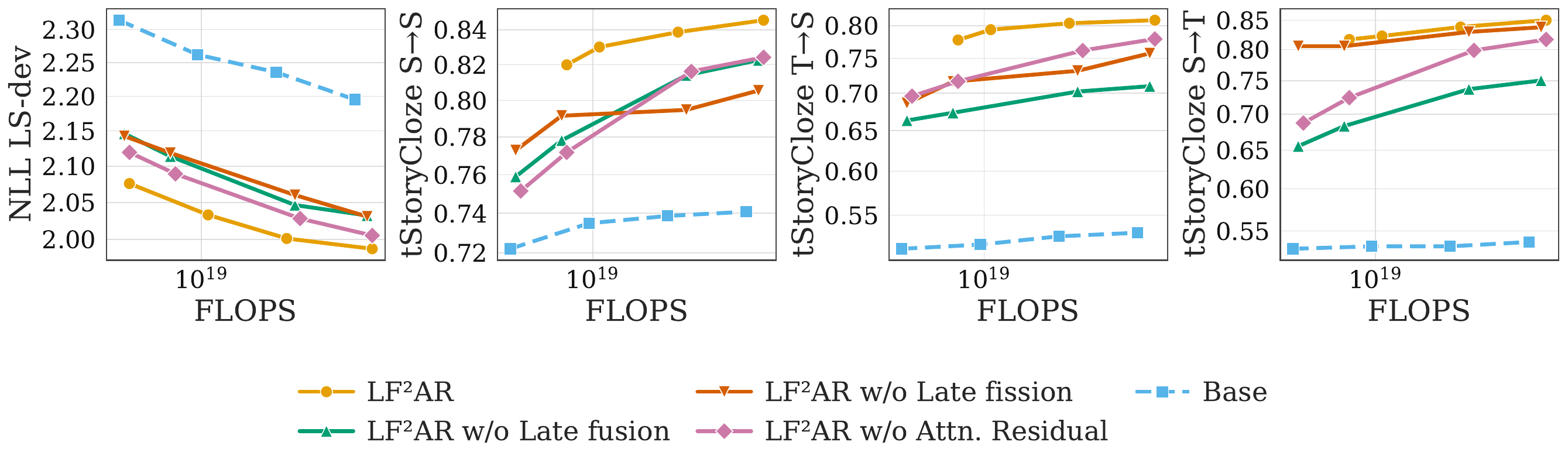}}
\caption{Downstream benchmark performance for speech-adapted LMs across subtractive ablations from \emph{LF²AR}. Late fusion has the largest individual effect, while late fission and attention residuals provide additional gains, especially in cross-modal settings.}
\label{fig:speech_pixlm_ablation}
\end{center}
\vskip -0.2in
\end{figure}

\begin{figure}[h]
\begin{center}
\centerline{\includegraphics[width=\textwidth]{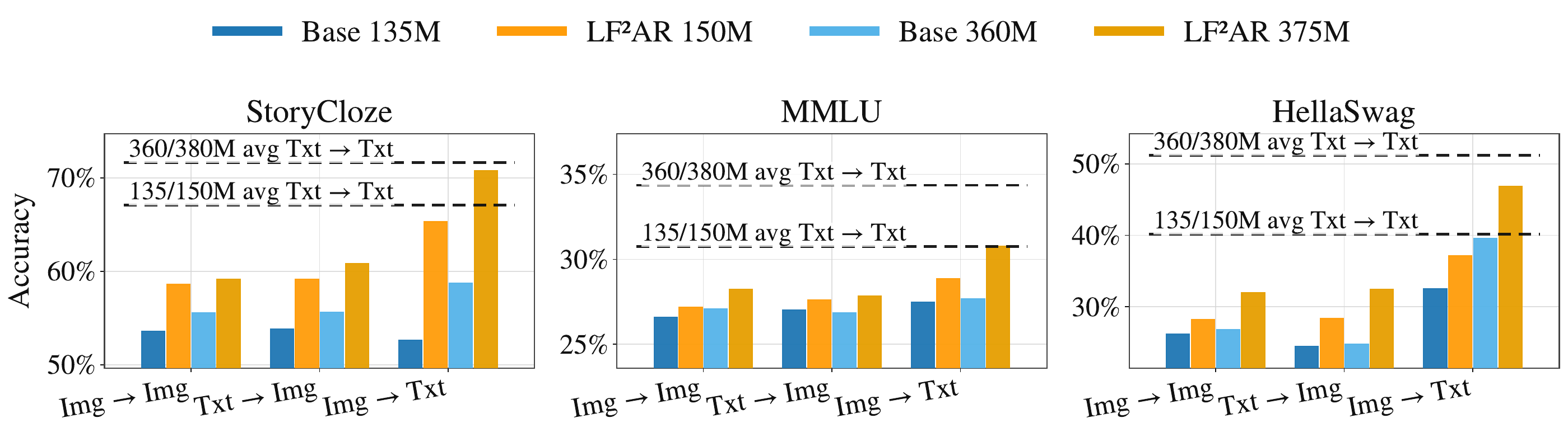}}
\caption{Scaling curves for pixel-adapted LMs. \emph{LF²AR} outperforms \emph{Base} at both scales, with the largest gains in Img→Txt, indicating that the architectural changes improve transfer efficiency beyond parameter scaling alone.}
\label{fig:scaling_perf_pixlm}
\end{center}
\vskip -0.2in
\end{figure}

\begin{table*}[h]
\centering
\setlength{\tabcolsep}{4pt} 
\resizebox{\textwidth}{!}{%
    \begin{tabular}{@{}lccccccccccc@{}}
        \toprule
        \multirow{2}{*}{Model} & \multirow{2}{*}{\shortstack{Params.}} & \multirow{2}{*}{\shortstack{Tokens}} & \multicolumn{4}{c}{tStoryCloze} & \multicolumn{4}{c}{sStoryCloze} & \multirow{1}{*}{MMLU} \\
        \cmidrule(lr){4-7} \cmidrule(lr){8-11} \cmidrule(lr){12-12}
         &  &  & \texttt{T} & \texttt{S} & \texttt{T$\to$S} & \texttt{S$\to$T} & \texttt{T} & \texttt{S} & \texttt{T$\to$S} & \texttt{S$\to$T} & \texttt{T} \small{(\texttt{adapted}/\texttt{base})} \\
        \midrule
        SpiritLM-Base-7B & 7B & $\sim$175B & 95.8 & \textbf{\underline{90.5}} & 78.6 & \textbf{\underline{94.3}} & 74.0 & \textbf{\underline{66.3}} & \textbf{\underline{64.7}} & \textbf{71.7} & 37.7 / 39.0 \\
        GLM-4-Voice 1.5B & 1.5B & 1T & --- & 77.5 & 81.4 & 90.1 & --- & 55.4 & 58.6 & 64.0 & --- \\
        GLM-4-Voice 9B & 9B & 1T & --- & 83.0 & \textbf{\underline{85.0}} & \textbf{93.6} & --- & \textbf{62.4} & 63.2 & \textbf{\underline{76.3}} & --- \\
        \midrule
        \textsc{LF²AR}-150M & 150M & 16B & 88.4 & 82.0 & 75.2 & 81.0 & 64.1 & 55.0 & 58.8 & 58.4 & 30.0 / 30.2 \\
        \textsc{LF²AR}-400M & 400M & 16B & 91.3 & 84.6 & 80.9 & 85.0 & 68.4 & 57.5 & 62.3 & 62.1 & 34.0 / 34.2 \\
        \textsc{LF²AR}-2B & 2B & 32B & 92.6 & \textbf{87.6} & \textbf{84.3} & 87.1 & 73.6 & 61.4 & \textbf{64.2} & 64.2 & \textbf{\underline{40.1 / 40.0}}\\
        \bottomrule
    \end{tabular}
}
\caption{Comparison against literature baselines on downstream evaluations. \textbf{\underline{Best results}} are bold and underlined; \textbf{second best} are bold. For MMLU, “best” denotes the model with the smallest post-adaptation degradation relative to its text-only base model.}
\label{tab:benchmark}
\end{table*}

\clearpage

\section{Transformer Background}
\label{app:transformer-background}

We briefly fix terminology for the transformer architecture used throughout the paper.
Our models are causal decoder-only transformers that process a sequence of discrete tokens
from a vocabulary $\mathcal{V}$ autoregressively through a stack of layers. Let
$c_i^{(l)} \in \mathbb{R}^d$ denote the representation at sequence position $i$ after layer
$l$. We refer to the sequence $\{c_i^{(l)}\}_{l=0}^{L}$ as the \emph{residual stream} at
position $i$. Each decoder layer updates the current residual-stream state by adding the output of its
submodules to the running representation inherited from previous layers. Ignoring
normalization details, a layer can be written as
\begin{align}
\hat{c}_i^{(l)} &= c_i^{(l-1)} + \mathrm{Attn}^{(l)}(c_{\leq i}^{(l-1)}), \\
c_i^{(l)} &= \hat{c}_i^{(l)} + \mathrm{MLP}^{(l)}(\hat{c}_i^{(l)}).
\end{align}
Thus, rather than replacing the previous representation, each layer incrementally refines a
shared state. This residual-stream view is useful because intermediate layers can be
understood as exposing different levels of abstraction of the same token position. In an autoregressive language model, the top-layer representation $c_i^{(L)}$ is projected
to logits over the next token, defining a distribution over $\mathcal{V}$ conditioned on the
preceding context. In this paper, we use the terminology \emph{backbone} to refer to this
stack of pretrained transformer layers.

\section*{Author contributions}

Santiago proposed and led the project, defining the hypotheses, model, and experimental design, developed most of the training, modeling, evaluation, and analysis codebase, and led the writing of the manuscript. Adel contributed to speech data preprocessing, synthesis, and evaluation, while also contributing model-design insights. Ricard provided supervision throughout and contributed to the data pipeline, experimental codebase, and analysis. Yanis assisted with experiment implementation, execution, and discussions. Phil contributed to the manuscript and discussions, as well as to Adel's supervision.. The other authors contributed to early-stage discussions and supervision during the JSALT 2024 workshop.

\end{document}

%% file: img/lf2ar_architecture_inline_tikz.tex

\begin{figure*}[t]
  \centering

  \definecolor{Peach}{HTML}{FEDCBD}
  \definecolor{PinkTop}{HTML}{FDECE9}
  \definecolor{PinkBottom}{HTML}{F8D8D6}
  \definecolor{YellowTop}{HTML}{FFF6C9}
  \definecolor{YellowBottom}{HTML}{FBE59B}

  \definecolor{BackboneFill}{HTML}{F3F0FB}
  \definecolor{BackboneStroke}{HTML}{3955AE}

  \definecolor{AttnLight}{HTML}{EDF4FF}
  \definecolor{AttnDark}{HTML}{082B83}
  \definecolor{AttnStroke}{HTML}{2858B8}

  \definecolor{TokenYellow}{HTML}{FFF1AE}
  \definecolor{TextToken}{HTML}{79C9C3}
  \definecolor{WordBlue}{HTML}{142BFF}

  \definecolor{WaveGray}{HTML}{888888}
  \definecolor{WaveLight}{HTML}{CCCCCC}

  \resizebox{\textwidth}{!}{%
\begin{tikzpicture}[
  x=1cm,
  y=1cm,
  line cap=round,
  line join=round
]

  \path[use as bounding box] (-0.75,0) rectangle (17.90,8.64);
  \fill[white] (-0.75,0) rectangle (17.90,8.64);


  \tikzset{
    titlefont/.style={
      font=\sffamily\itshape\fontsize{11.0}{12.4}\selectfont
    },
    labelfont/.style={
      font=\sffamily\itshape\fontsize{8.8}{10}\selectfont
    },
    rowfont/.style={
      font=\sffamily\itshape\fontsize{8.2}{9.2}\selectfont
    },
    legendheadfont/.style={
      font=\sffamily\itshape\fontsize{7.5}{8.4}\selectfont
    },
    legenditemfont/.style={
      font=\sffamily\itshape\fontsize{6.6}{7.4}\selectfont
    },
    wordfont/.style={
      font=\sffamily\itshape\fontsize{7.4}{8.4}\selectfont,
      text=WordBlue
    },
    backbone/.style={
      rounded corners=1.6mm,
      draw=BackboneStroke,
      line width=.65pt,
      fill=BackboneFill
    }
  }

  \def\xmin{1.46}
  \def\xmax{15.38}

  \def\xmodmin{6.05}
  \def\xmodmax{15.40}
  \def\xmodmid{10.725}

  %

  \newcommand{\attncircle}[3]{%
    \pgfmathtruncatemacro{\attnpct}{round(100*(#3))}
    \draw[
      draw=AttnStroke,
      line width=.55pt,
      fill={AttnDark!\attnpct!AttnLight}
    ]
      (#1,#2) circle[radius=.095];
  }

  %

  \newcommand{\texttokenrow}[2]{%
    \foreach \xc in {2.78,3.55,4.32,5.09,5.86}{
      \draw[
        fill=#2,
        draw=black,
        line width=.55pt
      ]
        ({\xc-0.115},#1) rectangle ++(0.23,0.23);
    }
  }


  \newcommand{\speechtokenrow}[2]{%

    \foreach \i in {11,...,19}{
      \pgfmathsetmacro{\xx}{1.57+0.42*\i}
      \draw[
        fill=#2,
        draw=black,
        line width=.55pt
      ]
        (\xx,#1) rectangle ++(0.23,0.23);
    }

    \node[
      font=\fontsize{11}{11}\selectfont
    ] at (10.27,#1+0.115) {$\cdots$};

    \foreach \i in {0,...,10}{
      \pgfmathsetmacro{\xx}{10.89+0.42*\i}
      \draw[
        fill=#2,
        draw=black,
        line width=.55pt
      ]
        (\xx,#1) rectangle ++(0.23,0.23);
    }
  }

  %

  \newcommand{\textbackbonerow}[1]{%
    \foreach \xc in {2.78,3.55,4.32,5.09,5.86}{
      \draw[
        draw=BackboneStroke,
        line width=.55pt,
        fill=BackboneFill
      ]
        (\xc,#1) circle[radius=.095];
    }
  }

  %
  %
  %
  %
  %

  \newcommand{\speechattentionrow}[3]{%

    \def\mode{#3}
    \def\topmode{t}


    \foreach \i in {9,...,17}{
      \pgfmathsetmacro{\xx}{2.73+0.41*\i}
      \pgfmathsetmacro{\aval}{#2}

      \ifnum\i=11
        \ifx\mode\topmode
          \pgfmathsetmacro{\aval}{1.0}
        \else
          \pgfmathsetmacro{\aval}{0.05}
        \fi
      \fi

      \ifnum\i=14
        \ifx\mode\topmode
          \pgfmathsetmacro{\aval}{1.0}
        \else
          \pgfmathsetmacro{\aval}{0.05}
        \fi
      \fi

      \ifnum\i=17
        \ifx\mode\topmode
          \pgfmathsetmacro{\aval}{1.0}
        \else
          \pgfmathsetmacro{\aval}{0.05}
        \fi
      \fi

      \attncircle{\xx}{#1}{\aval}
    }

    \node[
      font=\fontsize{11}{11}\selectfont
    ] at (10.17,#1+0.01) {$\cdots$};


    \foreach \i in {0,...,11}{
      \pgfmathsetmacro{\xx}{10.70+0.40*\i}
      \pgfmathsetmacro{\aval}{#2}

      \ifnum\i=1
        \ifx\mode\topmode
          \pgfmathsetmacro{\aval}{1.0}
        \else
          \pgfmathsetmacro{\aval}{0.05}
        \fi
      \fi

      \ifnum\i=6
        \ifx\mode\topmode
          \pgfmathsetmacro{\aval}{1.0}
        \else
          \pgfmathsetmacro{\aval}{0.05}
        \fi
      \fi

      \attncircle{\xx}{#1}{\aval}
    }
  }

  %

  \speechtokenrow{8.05}{Peach}


  \shade[
    rounded corners=1.8mm,
    top color=PinkTop,
    bottom color=PinkBottom
  ]
    (\xmodmin,7.22) rectangle (\xmodmax,7.86);

  \draw[
    rounded corners=1.8mm,
    line width=.7pt,
    black
  ]
    (\xmodmin,7.22) rectangle (\xmodmax,7.86);

  \node[titlefont] at (\xmodmid,7.54)
    {Refinement Layers};


  \draw[backbone]
    (\xmin,5.82) rectangle (\xmax,6.39);

  \draw[backbone]
    (\xmin,5.13) rectangle (\xmax,5.70);

  \draw[backbone]
    (\xmin,4.44) rectangle (\xmax,5.01);

  \draw[backbone]
    (\xmin,3.75) rectangle (\xmax,4.32);

  %

  \texttokenrow{6.53}{TextToken}

  %

  \foreach \i in {9,...,17}{
    \pgfmathsetmacro{\xx}{2.73+0.41*\i}

    \draw[
      AttnStroke,
      line width=.55pt,
      densely dashed,
      -{Latex[length=1.5mm,width=1.0mm]}
    ]
      (\xx,4.035) -- (\xx,7.16);
  }

  \foreach \i in {0,...,11}{
    \pgfmathsetmacro{\xx}{10.70+0.40*\i}

    \draw[
      AttnStroke,
      line width=.55pt,
      densely dashed,
      -{Latex[length=1.5mm,width=1.0mm]}
    ]
      (\xx,4.035) -- (\xx,7.16);
  }


  \node[
    rowfont,
    anchor=east
  ] at (2.59,6.105)
    {Layer L};

  \foreach \yy in {5.505,5.415,5.325}{
    \fill[black] (2.05,\yy) circle[radius=.022];
  }

  \node[
    rowfont,
    anchor=east
  ] at (2.59,4.725)
    {Layer 2};

  \node[
    rowfont,
    anchor=east
  ] at (2.59,4.035)
    {Layer 1};


  \textbackbonerow{6.105}
  \textbackbonerow{5.415}
  \textbackbonerow{4.725}
  \textbackbonerow{4.035}

  %
  %

  \speechattentionrow{6.105}{0.05}{t}
  \speechattentionrow{5.415}{0.20}{n}
  \speechattentionrow{4.725}{0.35}{n}
  \speechattentionrow{4.035}{0.50}{n}


  \node[
    labelfont,
    align=center,
    anchor=east
  ] at (1.12,4.73)
    {Backbone\\Layers};

  \draw[
    decorate,
    decoration={
      brace,
      amplitude=4pt,
      mirror
    },
    line width=.7pt
  ]
    (1.37,6.37) -- (1.37,3.76);



\draw[
  rounded corners=1.8mm,
  line width=.65pt,
  black,
  fill=white
]
  (15.54,3.48) rectangle (17.78,5.50);

\node[
  legendheadfont,
  align=center
] at (16.66,5.06)
  {Attn. Residual\\Weight};

\attncircle{15.90}{4.64}{0.0}

\node[
  legenditemfont,
  anchor=west
] at (16.25,4.64)
  {Low};

\attncircle{15.90}{4.23}{0.5}

\node[
  legenditemfont,
  anchor=west
] at (16.25,4.23)
  {Mid};

\attncircle{15.90}{3.82}{1.0}

\node[
  legenditemfont,
  anchor=west
] at (16.25,3.82)
  {High};


\draw[
  rounded corners=1.8mm,
  line width=.65pt,
  black,
  fill=white
]
  (15.54,2.12) rectangle (17.78,3.40);

\node[
  legendheadfont,
  align=center
] at (16.66,3.17)
  {Token Modality};

\draw[
  fill=Peach,
  draw=black,
  line width=.55pt
]
  (15.80,2.72) rectangle ++(0.20,0.20);

\node[
  legenditemfont,
  anchor=west
] at (16.25,2.82)
  {Speech};

\draw[
  fill=TextToken,
  draw=black,
  line width=.55pt
]
  (15.80,2.31) rectangle ++(0.20,0.20);

\node[
  legenditemfont,
  anchor=west
] at (16.25,2.41)
  {Text};

  %

  \texttokenrow{3.29}{TextToken}

  \speechtokenrow{3.29}{TokenYellow}


  \shade[
    rounded corners=1.8mm,
    top color=YellowTop,
    bottom color=YellowBottom
  ]
    (\xmodmin,2.52) rectangle (\xmodmax,3.08);

  \draw[
    rounded corners=1.8mm,
    line width=.7pt,
    black
  ]
    (\xmodmin,2.52) rectangle (\xmodmax,3.08);

  \node[titlefont] at (\xmodmid,2.80)
    {Composition Layers};

  %

  \speechtokenrow{2.14}{Peach}

  %

  \pgfmathsetseed{260806}

  \begin{scope}

    \clip (6.12,1.02) rectangle (15.40,1.96);

    \foreach \i in {0,...,720}{
      \pgfmathsetmacro{\t}{\i/720}
      \pgfmathsetmacro{\xw}{1.45+13.90*\t}

      \pgfmathsetmacro{\env}{
        0.040
        +0.72*exp(-pow((\t-0.025)/0.010,2))
        +0.46*exp(-pow((\t-0.052)/0.012,2))
        +0.62*exp(-pow((\t-0.083)/0.014,2))
        +0.34*exp(-pow((\t-0.112)/0.009,2))
        +0.88*exp(-pow((\t-0.155)/0.022,2))
        +0.45*exp(-pow((\t-0.205)/0.013,2))
        +0.78*exp(-pow((\t-0.245)/0.019,2))
        +0.38*exp(-pow((\t-0.292)/0.012,2))
        +0.92*exp(-pow((\t-0.345)/0.025,2))
        +0.52*exp(-pow((\t-0.400)/0.015,2))
        +0.70*exp(-pow((\t-0.447)/0.021,2))
        +0.42*exp(-pow((\t-0.500)/0.013,2))
        +0.82*exp(-pow((\t-0.555)/0.024,2))
        +0.48*exp(-pow((\t-0.615)/0.016,2))
        +0.76*exp(-pow((\t-0.662)/0.020,2))
        +0.95*exp(-pow((\t-0.725)/0.027,2))
        +0.43*exp(-pow((\t-0.782)/0.013,2))
        +0.68*exp(-pow((\t-0.825)/0.019,2))
        +0.86*exp(-pow((\t-0.885)/0.025,2))
        +0.50*exp(-pow((\t-0.940)/0.016,2))
        +0.66*exp(-pow((\t-0.975)/0.012,2))
      }

      \pgfmathsetmacro{\r}{0.60+0.55*rnd}

      \pgfmathsetmacro{\carrier}{
        0.24
        +0.46*abs(sin(14800*\t+52*sin(1040*\t)))
        +0.30*abs(sin(11900*\t+31*sin(1730*\t)))
      }

      \pgfmathsetmacro{\amp}{
        min(0.40,0.39*\env*\carrier*\r+0.016*rnd)
      }

      \draw[
        WaveLight,
        opacity=.78,
        line width=.28pt
      ]
        (\xw,{1.49-1.05*\amp})
        --
        (\xw,{1.49+1.05*\amp});
    }

    \foreach \i in {0,...,1150}{
      \pgfmathsetmacro{\t}{\i/1150}
      \pgfmathsetmacro{\xw}{1.45+13.90*\t}

      \pgfmathsetmacro{\env}{
        0.035
        +0.72*exp(-pow((\t-0.025)/0.010,2))
        +0.46*exp(-pow((\t-0.052)/0.012,2))
        +0.62*exp(-pow((\t-0.083)/0.014,2))
        +0.34*exp(-pow((\t-0.112)/0.009,2))
        +0.88*exp(-pow((\t-0.155)/0.022,2))
        +0.45*exp(-pow((\t-0.205)/0.013,2))
        +0.78*exp(-pow((\t-0.245)/0.019,2))
        +0.38*exp(-pow((\t-0.292)/0.012,2))
        +0.92*exp(-pow((\t-0.345)/0.025,2))
        +0.52*exp(-pow((\t-0.400)/0.015,2))
        +0.70*exp(-pow((\t-0.447)/0.021,2))
        +0.42*exp(-pow((\t-0.500)/0.013,2))
        +0.82*exp(-pow((\t-0.555)/0.024,2))
        +0.48*exp(-pow((\t-0.615)/0.016,2))
        +0.76*exp(-pow((\t-0.662)/0.020,2))
        +0.95*exp(-pow((\t-0.725)/0.027,2))
        +0.43*exp(-pow((\t-0.782)/0.013,2))
        +0.68*exp(-pow((\t-0.825)/0.019,2))
        +0.86*exp(-pow((\t-0.885)/0.025,2))
        +0.50*exp(-pow((\t-0.940)/0.016,2))
        +0.66*exp(-pow((\t-0.975)/0.012,2))
      }

      \pgfmathsetmacro{\rA}{0.50+0.72*rnd}
      \pgfmathsetmacro{\rB}{0.91+0.18*rnd}

      \pgfmathsetmacro{\carrier}{
        0.16
        +0.50*abs(sin(15100*\t+61*sin(1260*\t)))
        +0.24*abs(sin(12700*\t+37*sin(2210*\t)))
        +0.10*abs(sin(9300*\t))
      }

      \pgfmathsetmacro{\amp}{
        min(0.36,0.34*\env*\carrier*\rA+0.009*rnd)
      }

      \draw[
        WaveGray,
        opacity=.92,
        line width=.20pt
      ]
        (\xw,{1.49-\amp/\rB})
        --
        (\xw,{1.49+\amp*\rB});
    }

  \end{scope}


  \node[wordfont] at (2.78,1.49) {Hierarchy};
  \node[wordfont] at (3.55,1.49) {is};
  \node[wordfont] at (4.32,1.49) {one};
  \node[wordfont] at (5.09,1.49) {of};
  \node[wordfont] at (5.86,1.49) {the};


  \newcommand{\wordseg}[3]{%
    \draw[
      WordBlue,
      line width=.65pt
    ]
      (#1,0.82) -- (#2,0.82);

    \draw[
      WordBlue,
      line width=.65pt
    ]
      (#1,0.75) -- (#1,0.89);

    \draw[
      WordBlue,
      line width=.65pt
    ]
      (#2,0.75) -- (#2,0.89);

    \node[wordfont] at ({(#1+#2)/2},0.50)
      {#3};
  }

  \wordseg{6.12}{7.19}{central}
  \wordseg{7.37}{8.47}{structural}
  \wordseg{8.65}{9.75}{schemes}
  \wordseg{9.92}{10.49}{that}
  \wordseg{10.66}{11.22}{the}
  \wordseg{11.38}{12.39}{architect}
  \wordseg{12.55}{13.10}{of}
  \wordseg{13.23}{14.48}{complexity}
  \wordseg{14.64}{15.38}{uses}

\end{tikzpicture}
  }%

  \caption{
    Late Fusion and Late Fission with Attention Residuals (LF\textsuperscript{2}AR), illustrated for a
    text--speech sequence. Text interfaces directly with the pretrained backbone,
    while speech representations pass through additional composition and refinement
    layers. An attention residual allows the refinement module to selectively access
    residual-stream states across backbone depths. The attention weights illustrate
    our hypothesized sparse use of depth.
  }
  \label{fig:lf2ar-architecture}
\end{figure*}